\documentclass[letterpaper]{article}

\usepackage[hyphens]{url}
\usepackage{graphicx}
\usepackage{natbib}
\usepackage{caption}
\usepackage{algorithm}
\usepackage{algorithmic}
\usepackage{booktabs}
\usepackage{amsmath}
\usepackage{amssymb}
\usepackage{multirow}
\usepackage[hidelinks]{hyperref}
\usepackage[margin=1in]{geometry}
\usepackage{authblk}

\newcommand{\dataset}{BurstGPT}

\title{
Diagnose Before You Compress:\\
Prediction-Independent Bottleneck Witness Refinement\\
for LLM Serving Traces
}

\author[1]{Liming Liu}
\author[1]{Chao Hu}
\author[2]{Mingfei Lu}
\author[3]{Cong Tan}
\author[1]{Yiwei Ge}
\author[4]{Chijin Zhou}
\author[1]{Yongjun Xie}
\author[5]{Runzhe Wang}
\author[5]{Xiaohai Shi}
\author[1]{Heyuan Shi}

\affil[1]{Central South University, Changsha, China}
\affil[2]{University of Technology Sydney, Sydney, Australia}
\affil[3]{Chongqing Normal University, Chongqing, China}
\affil[4]{East China Normal University, Shanghai, China}
\affil[5]{Alibaba Group, China}

\affil[]{
\small
\texttt{\{244512048,8206240605\}@csu.edu.cn},
\texttt{huchao@csu.edu.cn},
\texttt{xieyongjun@csu.edu.cn}\\
\texttt{hey.shi@foxmail.com},
\texttt{mingfei.lu@student.uts.edu.au}\\
\texttt{2024210516076@stu.cqnu.edu.cn},
\texttt{cjzhou@sei.ecnu.edu.cn}\\
\texttt{\{runzhe.wrz,xiaohai.sxh\}@alibaba-inc.com}
}

\date{}

\begin{document}

\maketitle

\pagestyle{empty}
\thispagestyle{empty}

\begin{abstract}
Production LLM serving generates millions of diverse requests,
making full-trace replay across serving configurations increasingly
expensive. Existing trace reduction methods mainly preserve workload
distributions or representative requests, but bottleneck-revealing
workloads may be rare and non-representative. Moreover, evidence for one
component cannot compensate for missing evidence in another, while using
predicted bottlenecks as target truth creates circular evaluation. These
limitations make it necessary to preserve evidence for every bottleneck
component rather than rely on workload representativeness alone.
We propose Bottleneck-Preserving Witnessing (BPW), a quality-constrained
framework for compact and diagnostically reliable LLM serving replay
suites. BPW first performs Workload Candidate Nomination using
response-blind workload features and closed source-side measurements.
This stage identifies workloads that may expose scheduler, prefill,
decode, or KV-cache bottlenecks. Coverage-Priority Sequence Construction
then organizes multi-component proposals as reusable hyperedges and
prioritizes weak and uncovered dimensions. Finally, Bottleneck Truth
Verification derives prediction-independent labels solely from direct
target-system measurements. The verified results determine the earliest
prefix satisfying the direct two-witness requirement for every component.
Experiments on BurstGPT, ServeGen, and Mooncake show that BPW reaches the
verified gate with a compact workload set and outperforms 16 policies,
achieving relative improvements of 2.3\% and 16.3\% in Mean prefix Macro-F1 and WBRC-AUC, respectively. Stage-resolved and sensitivity analyses confirm the distinct
contributions and local stability of its three stages. Our code is publicly
available at \url{https://github.com/llmllmllm/BPW}.
\end{abstract}

\section{Introduction}

\begin{figure}[t]
    \centering
    \includegraphics[width=1\columnwidth]{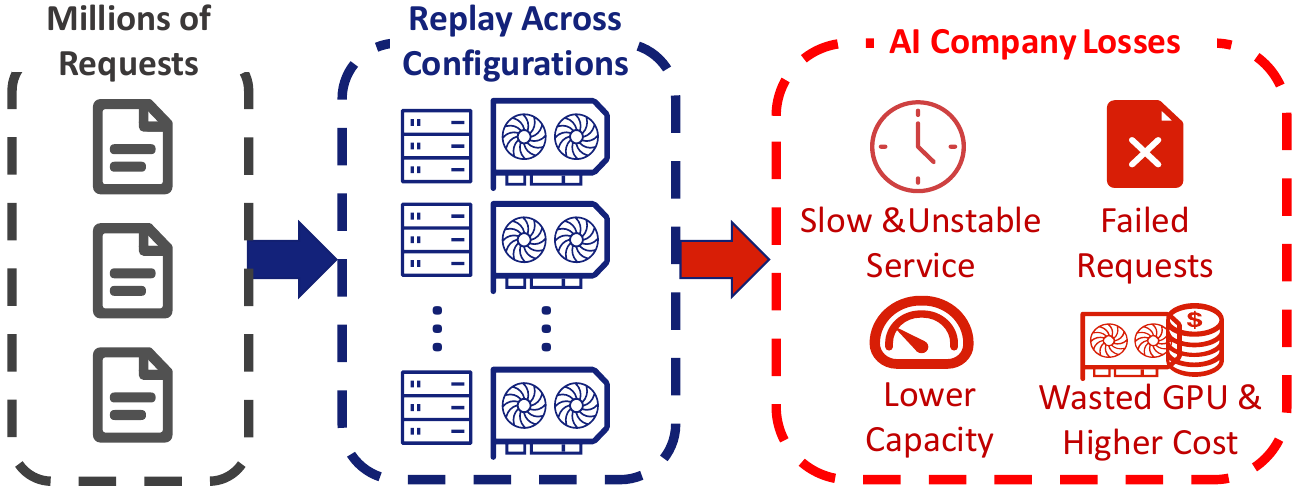}
    \caption{Practical costs of exhaustive replay and undetected bottlenecks in LLM serving.}
    \label{fig:intro-motivation}
\end{figure}

Reliable testing of LLM services requires identifying a small
set of critical workloads from millions of requests. As shown in
Figure~\ref{fig:intro-motivation}, exhaustive replay across serving
configurations is costly, while missed bottlenecks can degrade service
reliability, capacity, and GPU efficiency. Different workload patterns
can expose bottlenecks in request handling, computation, and memory
management that traffic may not reveal
\citep{sun2024llumnix,agrawal2024sarathi,
wang2025kvcache,gao2024cachedattention}. Such bottlenecks can cause slow
or failed requests, unstable service, reduced capacity, and higher
operating costs
\citep{sun2024llumnix,zhong2024distserve,qin2025mooncake}. Therefore,
improving test efficiency without losing critical bottleneck evidence is
a problem.

Existing work reduces evaluation cost by generating realistic workloads
\citep{xiang2026servegen}, selecting representative subsets that preserve
overall scores or model rankings
\citep{maiapolo2024tinybenchmarks,saranathan2025sublime,
perlitz2024efficient,li2025optimal}, or predicting untested outcomes from
a small number of evaluated cases
\citep{li2025aea,kossen2021active}. However, these objectives do not
ensure reliable component-level diagnosis. Diagnostically useful
workloads may be rare and may not appear representative; for example, an
infrequent workload can still expose a hidden bottleneck or failure mode
that is absent from dominant traffic patterns
\citep{xia2024fieldreplay,ma2025ndp}. One-time coverage is also insufficient, because later missed positives can
still weaken the evidence for some components. In addition, evidence for one
component cannot replace missing evidence for another. For example, many
Scheduler witnesses cannot compensate for the absence of a Prefill witness,
because the final requirement must hold for every component
\citep{zang2025adaptive}. Predicted outcomes also cannot be used as target
truth. For example, if the same source-side prediction both selects a workload
and labels it as a target bottleneck, the evaluation may simply confirm its
own prediction rather than the target system's actual behavior
\citep{nie2018adaptively}. Each bottleneck component needs its own evidence. Workload
representativeness alone is therefore insufficient. Consequently,
realistic workload generation, subset selection, and aggregate prediction
may still miss component bottlenecks.

To address these issues, we propose Bottleneck-Preserving Witnessing
(BPW), a unified framework for constructing compact and diagnostically
reliable LLM-serving replay suites. Within BPW,
Workload Candidate Nomination identifies rare
bottleneck-revealing workloads for verification;
Coverage-Priority Sequence Construction prioritizes evidence for
under-covered components under a limited budget; and Bottleneck Truth Verification
derives labels solely from direct target-system measurements, separating
selection from verification and avoiding adaptive-collection bias.
Experiments on three public traces show that BPW reaches the verified
four-component gate with few workloads and improves diagnostic quality.
Our contributions are threefold:
\begin{itemize}
    \item We formulate diagnostic trace reduction as quality-constrained
    bottleneck witnessing, prioritizing diagnostic quality over replay
    cost.

    \item We propose BPW, which combines multi-label hyperedges,
weakest-component protection, direct verification, and quality-constrained stopping.

\item We compare BPW with 16 policies and find that it improves Mean Prefix
Macro-F1 by 2.3 percent and WBRC-AUC by 16.3 percent on average.
\end{itemize}

\section{Background and Related Work}
\hspace*{\parindent}\textbf{LLM serving workloads and profiling.}\quad
Production traces and workload studies reveal the scale and diversity of
LLM services
\citep{ruan2026libra,yao2026opentela,gao2025weaver}.
% \dataset{} provides request timestamps, token counts, concurrency
% patterns, and failure information at large scale
% \citep{wang2025burstgpt}. Azure workloads, Mooncake, and ServeGen further
% describe conversational structure, prefix reuse, multi-turn requests, and
% production workload variation
% \citep{azure2024,qin2025mooncake,xiang2026servegen,
% xie2026strata}. Serving systems such as vLLM expose scheduling, latency,
% and memory behavior through runtime metrics
% \citep{kwon2023vllm,ruan2026libra,xie2026strata}. These studies provide realistic workloads and system
\dataset{} provides request timestamps, token counts, concurrency
patterns, and failure information at large scale
\citep{wang2025burstgpt}. Mooncake and ServeGen
describe conversational structure, prefix reuse, multi-turn requests, and
production workload variation
\citep{qin2025mooncake,xiang2026servegen,xie2026strata}.
Serving systems such as vLLM expose scheduling, latency,
and memory behavior through runtime metrics
\citep{kwon2023vllm,ruan2026libra,xie2026strata}.
These studies provide realistic workloads and system
measurements, but they do not provide independent
component-level bottleneck labels. In this work, a workload
becomes a bottleneck witness only when direct target-GPU
measurements confirm a change relative to a neutral anchor.

\textbf{Efficient evaluation and subset selection.} Prior efficient-evaluation methods select small subsets that preserve
full-set scores, model rankings, response patterns, or data distributions.
tinyBenchmarks estimates benchmark performance from a small set of items,
Active Evaluation Acquisition models dependencies among evaluation
items, and SubLIME selects subsets that preserve model rankings
\citep{maiapolo2024tinybenchmarks,li2025aea,saranathan2025sublime}.
Anchor-point and model-based evaluation methods further select informative
items to estimate full-set behavior with fewer evaluations
\citep{vivek2024anchorpoints,truong2025amortized}.
Predictive Kernel Herding selects workloads for accurate LLM
serving-performance prediction \citep{aydar2026pkh}, while coverage-based
and submodular objectives encourage diverse selections
\citep{bhargav2024submodular}. Software-testing methods also prioritize
configurations or reduce field-replay workloads
\citep{ma2025ndp,xia2024fieldreplay}.
However, preserving an average score, an overall ranking, or a workload
distribution does not guarantee that every component bottleneck is
directly exercised. Uncertainty-based selection may also spend budget on
samples that do not affect the final diagnostic decision
\citep{sloman2024nuisance}. Moreover, adaptive data collection must
remain separate from valid target inference
\citep{zrnic2024active}.

\section{Problem Formulation}
A small workload set may overlook important bottlenecks in the serving
system. We therefore study how to select a few informative workloads
that can still reveal and reliably verify all major bottlenecks. Formally,
let \(W\) be the target workload pool,
\(\mathcal{J}=\{\mathrm{sch},\mathrm{pre},\mathrm{dec},\mathrm{kv}\}\)
the registered bottleneck dimensions, and \(\mathcal{K}\) the serving
configurations. Executing workload \(i\in W\) across \(\mathcal{K}\)
costs \(C_i\) GPU seconds and produces direct target measurements
\(y_{ikj}\) for configuration \(k\) and bottleneck dimension \(j\).

Production inference workloads are often heterogeneous, bursty, and
difficult to predict
\citep{weng2022mlaas,khare2025superserve}. Because
bottleneck-revealing workloads may be rare and need not represent the
overall trace distribution, the objective is not to preserve a typical
subset of \(W\), but to prioritize workloads that are likely to expose
diagnostically important bottlenecks. A selector returns an ordered
sequence \(\pi=(\pi_1,\ldots,\pi_{|W|})\). After \(t\) executions, the
selected replay prefix is \(S_t=\{\pi_1,\ldots,\pi_t\}\).

We distinguish workload-level proposal quality from replay-suite quality.
Proposal quality measures how well workloads are prioritized, whereas
suite quality measures whether the selected prefix provides complete,
balanced, and stable bottleneck evidence. Similar to worst-group
evaluation, strong aggregate performance cannot compensate for poor
performance in an under-represented group
\citep{liu2021jtt}. Accordingly, evidence for one bottleneck dimension
cannot compensate for insufficient evidence in another, and a replay
prefix is feasible only after all registered diagnostic-quality
requirements are satisfied. Moreover, \(S_t\) is evaluated using
prediction-independent target truth. Proposal scores affect only workload
ordering and never enter the definition of target bottleneck labels,
preventing biased or circular evaluation
\citep{neel2018adaptive}. Workload count and GPU time are therefore
compared only among feasible prefixes.
% \paragraph{Difference from prior work.}
% Unlike methods that preserve workload representativeness or estimate
% full-trace performance, our task does not aim to reconstruct the complete
% response matrix, preserve an overall ranking, or match the dominant
% workload distribution. Instead, it seeks a compact replay prefix that
% directly witnesses every registered bottleneck dimension while maintaining
% high worst-dimension recall. One-time coverage is insufficient because
% later missed positives may still reduce the weakest component's recall.
% The selector must also remain separate from the target truth used for
% evaluation. The task therefore requires component-level evidence,
% quality-constrained stopping, and prediction-independent target
% verification.

\section{Methodology}

\begin{figure*}[t]
    \centering
    \includegraphics[width=\textwidth]{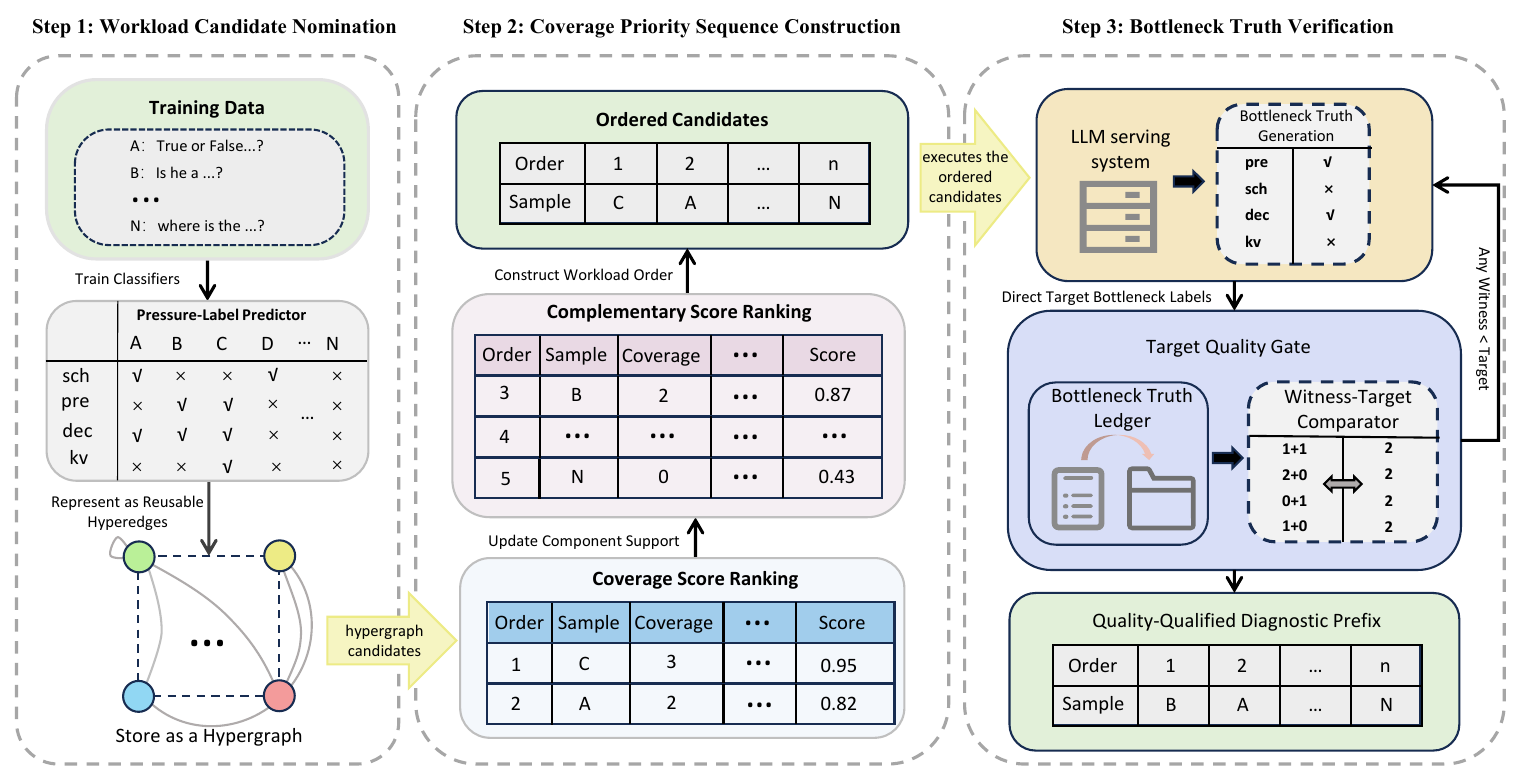}
    \caption{
Overview of BPW. Source-only predictors nominate multi-component bottleneck candidates, which are ordered by coverage priority and replayed on the target system. Direct target measurements independently verify bottleneck truth, and BPW returns the earliest prefix satisfying the diagnostic-quality and two-witness requirements.
}
    \label{fig:bpw_workflow}
\end{figure*}

We propose Bottleneck-Preserving Witnessing (BPW), a quality-constrained
framework for constructing compact yet diagnostically reliable workload
replay suites. As shown in Figure~\ref{fig:bpw_workflow}, BPW consists of
three stages. Workload Candidate Nomination takes the closed source
workload pool, response-blind pressure features, and source-side runtime
measurements as input, and outputs predicted multi-dimension bottleneck
labels together with a hypergraph-structured candidate suite.
Coverage-Priority Sequence Construction takes these candidate
hyperedges, their component-level proposal support, and execution-cost
information as input, and outputs an ordered workload sequence that
prioritizes weak and uncovered bottleneck dimensions.
Bottleneck Truth Verification executes the ordered candidates on
the target serving system and uses direct target measurements to produce
prediction-independent bottleneck labels, a bottleneck-evidence ledger,
and the earliest quality-qualified diagnostic prefix.

Let $W_{\mathrm{src}}$ and $W_{\mathrm{tar}}$ denote the closed source
workload pool and the target workload pool, respectively. Let
$\mathcal{J}=
\{\mathrm{sch},\mathrm{pre},\mathrm{dec},\mathrm{kv}\}$
index both the bottleneck dimensions and their
pressure coordinates, and let $\mathcal{K}$ denote the set of serving
configurations. Source measurements are used only to construct candidate
proposals for scheduler, prefill, decode, and KV-cache bottlenecks; they
do not define target truth. The resulting proposal information determines
which workloads are considered and how they are ordered, whereas direct
target measurements alone determine whether the proposed bottlenecks are
correct. BPW returns the earliest prefix that satisfies all predefined
diagnostic-quality constraints, and replay cost is considered only after
these constraints are met. BPW therefore integrates source-only candidate
proposal, component-aware workload ordering, independent truth verification,
and quality-constrained stopping.

\subsection{Workload Candidate Nomination}

For workload $i$, let $c_i$, $p_i$, $o_i$, and $r_i$ denote replay
concurrency, median input tokens, median output tokens, and prefix-reuse
ratio. We define four response-blind pressure coordinates as
$q_{i,\mathrm{sch}}=\log(1+c_i)$,
$q_{i,\mathrm{pre}}=\log(1+p_i)$,
$q_{i,\mathrm{dec}}=\log(1+o_i)$, and
$q_{i,\mathrm{kv}}=\log(1+v_i)$, where
$v_i=(p_i+o_i)c_i(1-0.75r_i)$.
Let $\mu_j^{\mathrm{src}}$ and $\sigma_j^{\mathrm{src}}$ denote the
frozen source mean and standard deviation of pressure dimension $j$.
The normalized coordinate is
$\widetilde q_{ij}=(q_{ij}-\mu_j^{\mathrm{src}})/
(\sigma_j^{\mathrm{src}}+\epsilon)$ for $j\in\mathcal J$, where
$\epsilon>0$ prevents numerical instability. We collect these
coordinates into
$\widetilde{\mathbf q}_i=
(\widetilde q_{i,\mathrm{sch}},
\widetilde q_{i,\mathrm{pre}},
\widetilde q_{i,\mathrm{dec}},
\widetilde q_{i,\mathrm{kv}})$.
The same frozen source statistics are applied to target workloads without
fitting on target responses.

A neutral anchor should have no unusually high pressure coordinate.
Accordingly, the source anchor is selected by lexicographically
minimizing its maximum pressure, total pressure, and workload index:
\begin{equation}
a_{\mathrm{src}}
=
\arg\min_{i\in W_{\mathrm{src}}}
\left[
\max_{j\in\mathcal J}\widetilde q_{ij},
\;
\sum_{j\in\mathcal J}\widetilde q_{ij},
\;
i
\right]_{\mathrm{lex}} .
\label{eq:source_anchor}
\end{equation}
where $[\cdot]_{\mathrm{lex}}$ denotes lexicographic comparison. The
target anchor $a_{\mathrm{tar}}$ is selected from $W_{\mathrm{tar}}$
using the same frozen normalization and tie-breaking rule.

We next use the source anchor to construct source-side bottleneck labels.
Let $\mathcal H$ denote the set of three chronological restarts. For
source environment $e$, restart $h\in\mathcal H$, workload $i$,
configuration $k\in\mathcal K$, and dimension $j\in\mathcal J$, let
$y^{e,h}_{ikj}$ denote the corresponding source response. Using the pooled
source responses, we construct an empirical cumulative distribution
function (ECDF) $F_{kj}$ for each combination of configuration $k$ and
dimension $j$, and compute
$z^{e,h}_{ikj}=F_{kj}(y^{e,h}_{ikj})$. The anchor-relative bottleneck
strength is
$d^{e,h}_{ij}=2\max\{\max_{k\in\mathcal K}
(z^{e,h}_{ikj}-z^{e,h}_{a_{\mathrm{src}}kj}),0\}$, where the maximum
selects the largest positive ECDF increase over the source anchor and
the factor of $2$ scales the score to the unit decision threshold
$d^{e,h}_{ij}\geq1$.

A source label is positive only when the condition is supported by the
majority of chronological restarts. Specifically,
$\ell^e_{ij}=\mathbb I[
\sum_{h\in\mathcal H}\mathbb I(d^{e,h}_{ij}\geq1)
\geq\lceil|\mathcal H|/2\rceil]$, where $\mathbb I[\cdot]$ denotes the
indicator function. Source labels are therefore defined only from source
measurements, and no predictor-side pressure threshold is used to
construct them.

For each bottleneck dimension, BPW trains a shallow Extra-Trees
classifier using the four normalized pressure coordinates. Model
selection is performed using source data only under
leave-one-candidate-out (LOCO) and leave-one-dataset-out (LODO)
validation. Let $f_j$ denote the resulting classifier for dimension $j$.
Its source-only proposal confidence is
$s_{ij}=f_j(\widetilde{\mathbf q}_i)$. A dimension is included in the
predicted bottleneck hyperedge $\widehat L_i$ when its proposal
confidence exceeds the pre-specified proposal threshold. The predicted
set $\widehat L_i$ is used only to propose and order workloads; neither
$s_{ij}$ nor $\widehat L_i$ enters the definition of target truth.

\subsection{Coverage-Priority Sequence Construction}

A workload may expose more than one bottleneck dimension. BPW therefore
represents $\widehat L_i$ as a reusable hyperedge rather than assigning
each workload to a single component, allowing one workload to provide
proposed evidence for several dimensions and reducing redundant replay
when multiple bottlenecks can be witnessed jointly. The candidate suite
$\mathcal C\subseteq W_{\mathrm{tar}}$ includes a neutral anchor,
workloads predicted to cover multiple bottleneck dimensions,
component-specific corroborating workloads, and a negative control. This
composition supports multi-component reuse while retaining independent
evidence for bottleneck dimensions and a reference for
false-positive behavior. Because source-side predictions may be less reliable for workloads far
from the closed source pressure support, BPW discounts their proposal
confidence using the distance from each target workload to the source
pressure support:
$\Delta_i=\min_{m\in W_{\mathrm{src}}}
\|\widetilde{\mathbf q}_i-\widetilde{\mathbf q}_m\|_2$.
Let $\omega:[0,\infty)\rightarrow(0,1]$ denote a non-increasing discount
function satisfying $\omega(0)=1$. The discounted proposal support is
$\bar s_{ij}=s_{ij}\omega(\Delta_i)$. This adjustment may lower the ordering priority of distant
workloads, but it does not remove them or define target truth. Such
workloads remain eligible for selection, and their bottlenecks are
confirmed only through direct target measurements.

Let $S_t=\{\pi_1,\ldots,\pi_t\}$ denote the prefix selected after $t$
steps of the frozen order $\pi$. The accumulated proposal support of
dimension $j$ is
$\rho_j(S_t)=\sum_{i\in S_t}\bar s_{ij}
\mathbb I(j\in\widehat L_i)$, and the weakest currently supported
dimension is $j_t^*=\arg\min_{j\in\mathcal J}\rho_j(S_t)$.
This weakest-component prioritization prevents strong evidence for one
component from compensating for weak evidence in another. To distinguish unsupported
dimensions from those that already have proposal evidence, let
$\mathcal U_t=\{j\in\mathcal J:\rho_j(S_t)=0\}$. For an unselected
candidate $i$, its dimension-specific marginal gain is
$g_t(i,j)=\bar s_{ij}\mathbb I(j\in\widehat L_i)/
(1+\rho_j(S_t))$. The denominator creates diminishing returns for
components that already have strong proposal support, while preserving
high gain for weak or uncovered components.

BPW combines weakest-component priority, uncovered-component gain,
and reusable hyperedge gain in the following diagnostic score:
\begin{equation}
\Psi_t(i)
=
\left[
g_t(i,j_t^*),
\sum_{j\in\mathcal U_t} g_t(i,j),
\sum_{j\in\mathcal J} g_t(i,j)
\right]_{\mathrm{lex}} .
\label{eq:diagnostic_score}
\end{equation}

\noindent
It then incorporates deterministic tie-breaking through

\begin{equation}
\Gamma_t(i)
=
\left[
\Psi_t(i),
-\kappa_i,
-i
\right]_{\mathrm{lex}} .
\label{eq:candidate_score}
\end{equation}
Here, $\kappa_i$ denotes the pre-specified measured or estimated execution
cost of workload $i$. Within $\Psi_t(i)$, the first entry prioritizes the
weakest component, the second rewards unsupported components, and the
third captures reusable multi-component gain, whereas in
$\Gamma_t(i)$, execution cost and workload index are used only as
deterministic tie-breakers. The next workload is selected as
$\pi_{t+1}=\arg\max_{i\in\mathcal C\setminus S_t}\Gamma_t(i)$, after
which the selected prefix is updated as
$S_{t+1}=S_t\cup\{\pi_{t+1}\}$ and the accumulated component support is
updated as $\rho_j(S_{t+1})=\rho_j(S_t)+
\bar s_{\pi_{t+1}j}\mathbb I(j\in\widehat L_{\pi_{t+1}})$.

\begin{algorithm}[t]
\caption{BPW Candidate Workload Ordering}
\label{alg:bpw}
\begin{algorithmic}[1]
\REQUIRE Candidate pool $\mathcal C$, predicted hyperedges
$\widehat L_i$, discounted support $\bar s_{ij}$,
execution cost $\kappa_i$, target anchor $a_{\mathrm{tar}}$
\STATE Initialize $\pi\leftarrow[a_{\mathrm{tar}}]$ and
$S_1\leftarrow\{a_{\mathrm{tar}}\}$
\STATE Compute $\rho_j(S_1)$ for all $j\in\mathcal J$
\STATE Initialize selection step $t\leftarrow1$
\WHILE{$\mathcal C\setminus S_t\neq\varnothing$}
    \STATE Identify $j_t^*\leftarrow
    \arg\min_{j\in\mathcal J}\rho_j(S_t)$
    \STATE Compute $\Gamma_t(i)$ for all
    $i\in\mathcal C\setminus S_t$
    \STATE Select
    $\pi_{t+1}\leftarrow
    \arg\max_{i\in\mathcal C\setminus S_t}\Gamma_t(i)$
    \STATE Update $S_{t+1}$ and $\rho_j(S_{t+1})$
    \STATE $t\leftarrow t+1$
\ENDWHILE
\STATE Append the registered negative control if it remains unselected
\RETURN Workload order $\pi$
\end{algorithmic}
\end{algorithm}

\subsection{Bottleneck Truth Verification}

After the candidate order, anchor, source ECDFs, and decision thresholds
are fixed, target workloads are executed relative to the target anchor
$a_{\mathrm{tar}}$. For restart $h\in\mathcal H$, let
$y^{\mathrm{tar},h}_{ikj}$ denote the direct target response of workload
$i$ under configuration $k$ and bottleneck dimension $j$. Using the
source ECDFs constructed before target verification, BPW computes
$z^{\mathrm{tar},h}_{ikj}
=F_{kj}(y^{\mathrm{tar},h}_{ikj})$.

The direct target bottleneck indicator for workload $i$, dimension $j$,
and restart $h$ is

\begin{equation}
b^{\mathrm{tar},h}_{ij}
=
\mathbb{I}
\left[
\max_{k\in\mathcal{K}}
\left(
z^{\mathrm{tar},h}_{ikj}
-
z^{\mathrm{tar},h}_{a^{\mathrm{tar}}kj}
\right)
\geq \delta_{\mathrm{tar}}
\right].
\label{eq:target_indicator}
\end{equation}
Here, $\delta_{\mathrm{tar}}\in(0,1]$ is a pre-specified
anchor-relative ECDF-gap threshold. It specifies the minimum increase
in the ECDF-transformed response of workload $i$ relative to the target
anchor $a^{\mathrm{tar}}$, under at least one serving configuration,
required for dimension $j$ to be identified as a target bottleneck.
A larger $\delta_{\mathrm{tar}}$ yields a more conservative bottleneck
criterion. The threshold is fixed before target evaluation to prevent
target-label leakage. The restart-specific target bottleneck set is
therefore
\[
L_i^{\mathrm{tar},h}
=
\{j\in\mathcal{J}:b^{\mathrm{tar},h}_{ij}=1\}.
\]
When a single target label across restarts is required, majority
agreement gives
\begin{equation}
L_i^{\mathrm{tar}}
=
\left\{
j\in\mathcal{J}:
\sum_{h\in\mathcal{H}} b^{\mathrm{tar},h}_{ij}
\geq
\left\lceil\frac{|\mathcal{H}|}{2}\right\rceil
\right\}.
\label{eq:target_majority}
\end{equation}
Target truth is defined only from direct target measurements under the
anchor-relative rule. It does not use $\widehat{L}_i$, $s_{ij}$,
$\bar{s}_{ij}$, a target-side pressure gate, or transferred source
support. A predicted bottleneck may therefore be absent from target
truth, while an unpredicted bottleneck may appear in it, keeping both
false triggers and missed bottlenecks visible.

Using these prediction-independent target labels, BPW evaluates each
prefix according to its predetermined workload order. It does not stop
at the first prefix that merely touches every bottleneck dimension,
because temporary full coverage may coexist with weak component-level
recall or unstable evidence across restarts. For prefix $S_t$, let
$N_{j,h}(S_t)=\sum_{i\in S_t}b^{\mathrm{tar},h}_{ij}$ denote the number
of distinct direct positive witnesses for component $j$ in restart $h$.
Robust component coverage is indicated by
$G_{\mathrm{rob}}(S_t)=
\mathbb I[\min_{h\in\mathcal H}\min_{j\in\mathcal J}
N_{j,h}(S_t)\geq2]$. Requiring two distinct witnesses ensures that every
component remains covered after deleting any single positive witness.

Let $M(S_t)$ denote Macro-F1,
$R_{\min}(S_t)=\min_{j\in\mathcal J}R_j(S_t)$ denote
worst-dimension recall, where $R_j(S_t)$ is the recall for dimension
$j$, and let $A(S_t)$ denote restart stability. We first define the
diagnostic-performance condition as
\begin{align}
\mathcal Q_{\mathrm{pred}}(S_t)
&=
\mathbb I\!\left[
M(S_t)\geq\eta_M
\land
R_{\min}(S_t)\geq\eta_R
\right].
\label{eq:predictive_quality}
\\
\intertext{We then define the evidence-reliability condition as}
\mathcal Q_{\mathrm{evid}}(S_t)
&=
\mathbb I\!\left[
G_{\mathrm{rob}}(S_t)=1
\land
A(S_t)\geq\eta_A
\right].
\label{eq:evidence_quality}
\end{align}

\noindent The diagnostic-quality gate is defined as
$\mathcal Q(S_t)=
\mathcal Q_{\mathrm{pred}}(S_t)
\mathcal Q_{\mathrm{evid}}(S_t)$.
Here, $\eta_M$, $\eta_R$, and $\eta_A$ denote the predefined thresholds
for Macro-F1, worst-dimension recall, and restart stability,
respectively. Thus, $\mathcal Q(S_t)=1$ indicates that both quality
conditions are satisfied. BPW returns the earliest feasible prefix
satisfying this gate:
$t^*=\min\{t\in\{1,\ldots,|\pi|\}:\mathcal Q(S_t)=1\}$.
If no prefix satisfies the gate, the selector is marked as diagnostically
infeasible rather than being assigned an artificially low replay cost.

Finally, let $\kappa_i^{\mathrm{tar}}$ denote the measured target
execution cost of workload $i$. The cost of a feasible order is
$C_{\mathrm{replay}}(\pi)=
\sum_{\ell=1}^{t^*}\kappa_{\pi_\ell}^{\mathrm{tar}}$, subject to
$\mathcal Q(S_{t^*})=1$. This sum accumulates the target execution costs
from the first workload through the earliest feasible prefix. Replay cost
is therefore compared only after diagnostic feasibility has been
established. Target responses determine only the target labels and the
quantities entering $\mathcal Q(S_t)$; they cannot modify the predictor,
source statistics, candidate pool, workload order, or stopping rule.

\section{Experimental Evaluation}
\label{sec:evaluation}
We evaluate BPW under diverse workload conditions and compare it against
representative trace reduction strategies. Our evaluation focuses on whether
BPW can achieve component-complete diagnosis with reduced replay cost, while
providing reliable evidence for its design choices and empirical robustness.

{\small
\begin{itemize}
    \item \textbf{RQ1: How effective and efficient is BPW?}
    \item \textbf{RQ2: What does BPW preserve beyond representativeness?}
    \item \textbf{RQ3: How do BPW's three stages contribute?}
    \item \textbf{RQ4: How stable are BPW's selections?}
\end{itemize}
}
\subsection{}{Datasets.}
Table~\ref{tab:datasets} summarizes three complementary public workload
sources. BurstGPT \cite{wang2025burstgpt} stresses arrival bursts and
scheduling; ServeGen \cite{xiang2026servegen} varies production-derived
request shapes; Mooncake \cite{qin2025mooncake} emphasizes prefix reuse and
KV-cache behavior in disaggregated serving. Together, they vary arrival rate, prompt length, generation length, and
prefix reuse, which are workload dimensions associated with Scheduler,
Prefill, Decode, and KV-cache bottlenecks. Source evidence is built
only from closed history. Each policy ranks the same finite target pool
response-blind and is scored on the same completed measurement matrix.

\begin{table}[!ht]
\vspace{-8pt}
\centering
\scriptsize
\setlength{\tabcolsep}{1.5pt}

\begin{tabular*}{\columnwidth}{
  @{\extracolsep{\fill}}lccccc@{}
}
\toprule
Trace
& \shortstack{Total\\Requests}
& \shortstack{Avg. Input\\Tokens}
& \shortstack{Avg. Output\\Tokens}
& \shortstack{Avg. Arrival\\Rate}
& \shortstack{Avg. Prefix\\Reuse} \\
\midrule
BurstGPT
  & 5.29M
  & 768.4
  & 502.1
  & 0.51
  & 0\% \\
ServeGen
  & 3.54B
  & 718.1
  & 332.2
  & 341.4
  & 42.8\% \\
Mooncake
  & 35.6K
  & 619.5
  & 303.6
  & 9.89
  & 25.7\% \\
\bottomrule
\end{tabular*}

\vspace{1pt}
% \caption{Source-trace scale and replay-workload characteristics.
% Requests and arrival rates describe source traces; token lengths
% and prefix reuse are averaged over replay workloads.}
\caption{Source-trace and replay-workload characteristics.}
\label{tab:datasets}
\vspace{-4pt}
\end{table}

\textbf{Parameters and Environment.}
Experiments use an NVIDIA Tesla V100S 32\,GB (driver 535.309.01),
Qwen2-1.5B-Instruct, vLLM 0.6.6.post1, PyTorch 2.5.1/CUDA 12.4, and an AMD
Ryzen 5 5600G host with 14\,GiB RAM. The formal GPU is capped at 100\,W with
300/1107\,MHz application graphics/memory clocks. We replay two vLLM
configurations: memory utilization 0.50 with at most 8 sequences, and 0.75
with at most 16; both use maximum model length and batched-token budget 4096.
Each cell contains 24 deterministic requests (temperature 0, EOS ignored) and
is repeated after three chronological server restarts.

\begin{table*}[!t]
\centering
\scriptsize
\setlength{\tabcolsep}{2.2pt}
\renewcommand{\arraystretch}{0.95}

\begin{tabular*}{0.99\textwidth}{@{\extracolsep{\fill}}lcccccc}
\toprule

& \multicolumn{2}{c}{BurstGPT}
& \multicolumn{2}{c}{ServeGen}
& \multicolumn{2}{c}{Mooncake}\\

\cmidrule(lr){2-3}
\cmidrule(lr){4-5}
\cmidrule(lr){6-7}

Method
& \shortstack{Mean prefix\\Macro-F1}
& \shortstack{WBRC\\AUC}
& \shortstack{Mean prefix\\Macro-F1}
& \shortstack{WBRC\\AUC}
& \shortstack{Mean prefix\\Macro-F1}
& \shortstack{WBRC\\AUC}\\

\midrule

Length (control)
& 0.6757 & 0.0281
& 0.8651 & 0.8335
& 0.8197 & 0.6020\\

Concurrency (control)
& 0.7864 & 0.5653
& 0.7552 & 0.7567
& 0.7961 & 0.7534\\

Mode-$k$ (control)
& 0.6653 & 0.2073
& 0.7697 & 0.6316
& 0.8470 & 0.8501\\

Pressure $k$-center
& 0.8193 & 0.6568
& 0.8436 & 0.6714
& 0.8523 & 0.8578\\

D-optimal
& 0.8301 & 0.5795
& 0.8594 & 0.7491
& 0.8542 & 0.7843\\

Facility location
& 0.6558 & 0.6955
& 0.7767 & 0.6741
& 0.8159 & 0.7646\\

Weighted source cover
& 0.6615 & 0.5848
& 0.8561 & 0.7902
& 0.8710 & 0.8004\\

PKH-inspired (EuroMLSys'26)
& 0.7513 & 0.6343
& 0.8593 & 0.7745
& 0.8731 & 0.8033\\

GP information gain
& 0.5604 & 0.6815
& 0.6977 & 0.6639
& 0.6745 & 0.6342\\

NDP-inspired (ICSE'25)
& 0.6407 & 0.4810
& 0.7424 & 0.7684
& 0.7939 & 0.7680\\

AEA-inspired (ICML'25)
& 0.6449 & 0.6828
& 0.7420 & 0.7554
& 0.8300 & 0.7920\\

FieldReplay-inspired (TSE'24)
& 0.6665 & 0.6343
& 0.7688 & 0.7340
& 0.8198 & 0.7734\\

SIS-inspired (L4DC'24)
& 0.7401 & 0.7258
& 0.8561 & 0.7902
& 0.8710 & 0.8004\\

tinyBench-inspired (ICML'24)
& 0.6665 & 0.6279
& 0.7688 & 0.7340
& 0.8210 & 0.7665\\

SubLIME-inspired (ACL'25)
& 0.4862 & 0.3812
& 0.7621 & 0.7257
& 0.7373 & 0.7030\\

SMART-inspired (NAACL'25)
& 0.6615 & 0.6312
& 0.7899 & 0.7601
& 0.8299 & 0.7849\\

\midrule

\textbf{BPW (ours)}
& \textbf{0.8553} & \textbf{0.9266}
& \textbf{0.8790} & \textbf{0.8763}
& \textbf{0.8915} & \textbf{0.9002}\\

\bottomrule
\end{tabular*}

\vspace{2pt}
\caption{Unified diagnostic-quality comparison. BPW achieves the highest
observed Mean prefix Macro-F1 and WBRC-AUC among the 16 evaluated reference
policies on BurstGPT, ServeGen, and Mooncake.}
\label{tab:v24-main}

\end{table*}

\textbf{Baselines.}
The 16 reference policies cover traffic controls (Length, Concurrency,
and Mode-$k$), geometric/design selection (pressure $k$-center, D-optimal,
facility location, and weighted source cover), information acquisition
(PKH-inspired kernel herding and GP information gain), and recent
adaptations: NDP-inspired (ICSE~2025), AEA-inspired (ICML~2025),
FieldReplay-inspired (TSE~2024), SIS-inspired (L4DC~2024),
tinyBench-inspired (ICML~2024), SubLIME-inspired (ACL~2025), and
SMART-inspired (NAACL~2025)
\cite{aydar2026pkh,ma2025ndp,li2025aea,xia2024fieldreplay,
bhargav2024submodular,maia2024tinybenchmarks,
saranathan2025sublime,gupta2025smart}. Most target distribution,
ranking, or representative-subset preservation rather than multi-label
bottleneck discovery. ``Inspired'' denotes a response-blind ranking
adaptation, not author-code reproduction. All policies receive identical
candidates, source features, costs, truth labels, and stopping rules;
thus, the comparison is limited to adapted general-purpose policies.

\textbf{Metrics.}
The non-compensatory quality gate requires at least two distinct direct
positive workloads for each of Scheduler, Prefill, Decode, and KV-cache.
We report two quality metrics, both computed over the complete executed-prefix
trajectory. Mean prefix Macro-F1 macro-averages component-label correctness
over Scheduler, Prefill, Decode, and KV-cache. The area under the worst-bottleneck recall versus cost curve
(WBRC-AUC) integrates the weakest component recall over normalized
cumulative execution cost; it is not a classification ROC-AUC.
Higher is better for both metrics. Workload count is reported separately as
suite cost and never compensates for missing component evidence.

\textbf{Diagnostic Quality and Efficiency.}
Table~\ref{tab:v24-main} shows that BPW achieves the highest observed
Mean prefix Macro-F1 and WBRC-AUC among the evaluated policies on all
three traces. Against the strongest baseline, the respective gains are
0.0252 and 0.2009 on BurstGPT, 0.0139 and 0.0428 on ServeGen, and 0.0185
and 0.0424 on Mooncake. BPW achieves these trajectory-wide gains while
reaching the direct two-witness gate with few workloads. Overall, BPW
ranks first in every displayed metric across the three public traces
among 16 reference policies.

\begin{figure}[!htb]
  \centering
  \IfFileExists{Figures/bpw_v24_motivation.pdf}
    {\includegraphics[width=0.98\columnwidth]{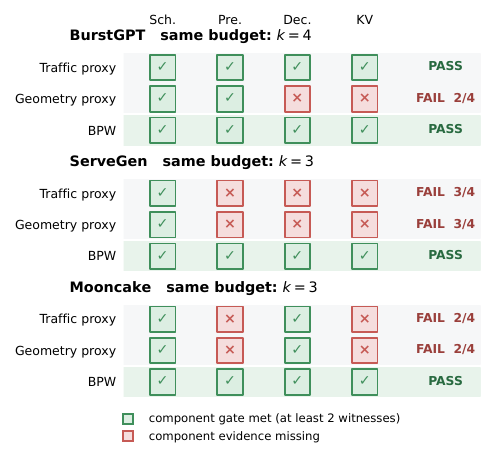}}
    {\fbox{\parbox[c][5.0cm][c]{0.94\columnwidth}{\centering
      Upload \texttt{Figures/bpw\_v24\_motivation.pdf}}}}
  \caption{Component-level evidence sufficiency under BPW’s equal budget. A green check indicates at least two direct positive witnesses for
the component, whereas a red cross indicates missing evidence. Pass requires
all four components and is consistent across three restarts.}
  \label{fig:motivation}
  \vspace{-14pt}
\end{figure}

\par\smallskip
\textbf{Complete Bottleneck Evidence Preservation.}
Figure~\ref{fig:motivation} compares component-level evidence sufficiency under equal workload budgets. It shows whether Scheduler,
Prefill, Decode, and KV-cache each receive the required two direct
witnesses. The two baselines preserve traffic
characteristics or workload geometry, allowing diagnostic sufficiency to
be compared independently of suite size. On BurstGPT, the geometry proxy
meets the requirement for only 2 of 4 components, whereas the traffic
proxy meets all four.

This agreement does not generalize across traces. On ServeGen, both
representative baselines miss three components, and on Mooncake, both
miss two. Under the same budgets, BPW is the only evaluated policy that
provides at least two direct witnesses for every component across all
three traces. These results show that preserving traffic characteristics
or workload geometry does not guarantee complete, non-compensatory
bottleneck evidence.

\begin{figure}[!t]
  \centering
  \IfFileExists{Figures/bpw_v24_ablation.pdf}
    {\includegraphics[width=0.98\columnwidth]{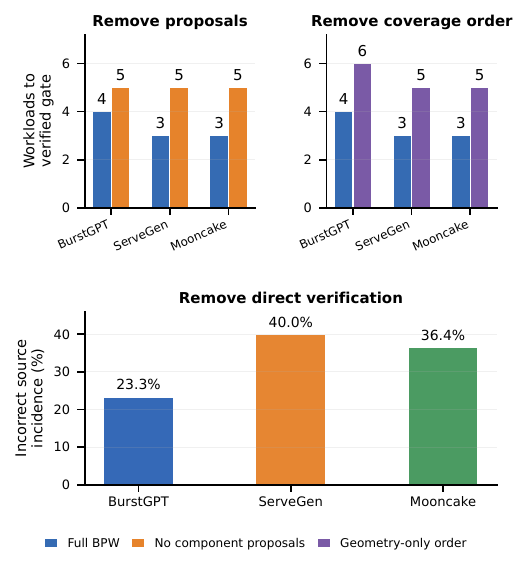}}
    {\fbox{\parbox[c][5.0cm][c]{0.94\columnwidth}{\centering
      Upload \texttt{Figures/bpw\_v24\_ablation.pdf}}}}
  \caption{Stage-resolved analysis on the same completed matrices. The
  upper panels remove component proposals or replace the
  coverage-prioritized order, while the lower panel reports incorrect
  component incidences without direct verification. Values are restart
  means; lower is better.}
  \label{fig:ablation}
  \vspace{-8pt}
\end{figure}

\textbf{Ablation Study.}
Figure~\ref{fig:ablation} analyzes the three stages of BPW: candidate
proposal, workload ordering, and direct verification.
Removing source component proposals increases the workloads required to
reach the direct two-witness gate across all three traces. Replacing the
state-aware coverage order with a geometry-only facility-location order
also increases the required workload count under the same candidate
pool, neutral anchor, and direct-truth protocol.

Source proposals remain informative, achieving high component Macro-F1 scores
across the three traces (0.884, 0.912, and 0.914 for BurstGPT, ServeGen, and
Mooncake, respectively). However, their predictions are not equivalent to
target truth: without direct verification, a non-negligible fraction of
candidates still contain incorrect component incidences (23.3\%, 40.0\%, and
36.4\% on BurstGPT, ServeGen, and Mooncake, respectively). These results
highlight the non-substitutable roles of the three stages. Proposals expose
candidate component structure, coverage ordering reduces the executions
required to satisfy the conjunctive gate, and direct measurements prevent
source hypotheses from being promoted as bottleneck truth. These controlled
stage interventions are performed on fixed measurement matrices and should be
interpreted as component contribution analyses rather than system-level causal
claims.

\par
\begin{figure}[!t]
  \centering
  \IfFileExists{Figures/bpw_v24_stability.pdf}
    {\includegraphics[width=0.92\columnwidth]
      {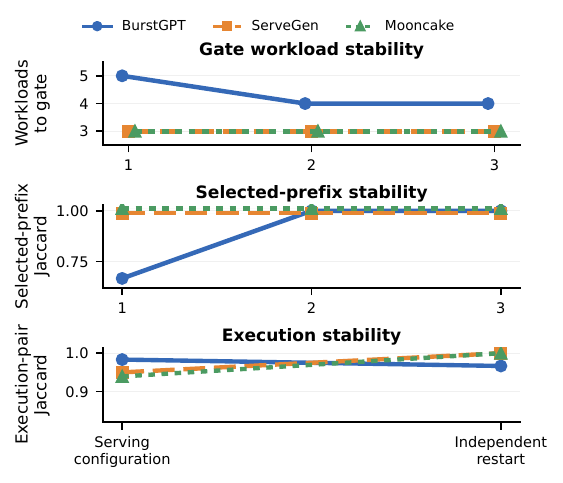}}
    {\fbox{\parbox[c][4.5cm][c]{0.94\columnwidth}{\centering
      Upload \texttt{Figures/bpw\_v24\_stability.pdf}}}}
  \vspace{-2pt}
  \caption{Stability of gate workload count, selected prefix, and execution
  pairs under parameter changes, serving configurations, and independent
  restarts. Values for ServeGen and Mooncake are slightly shifted for
  visibility; their selected-prefix Jaccard scores are both 1.000.}
  \label{fig:stability}
  \vspace{-10pt}
\end{figure}

\smallskip
\textbf{Stability and Evidence Boundary.}
Figure~\ref{fig:stability} evaluates gate-workload, selected-prefix, and
execution stability under parameter, configuration, and restart changes.
% \begin{figure}[!htb]
%   \centering
%   \IfFileExists{Figures/bpw_v24_stability.pdf}
%     {\includegraphics[width=0.92\columnwidth]{Figures/bpw_v24_stability.pdf}}
%     {\fbox{\parbox[c][4.5cm][c]{0.94\columnwidth}{\centering
%       Upload \texttt{Figures/bpw\_v24\_stability.pdf}}}}
%   \vspace{-2pt}
%   \caption{Stability of gate workload count, selected prefix, and execution
%   pairs under parameter changes, serving configurations, and independent
%   restarts. Values for ServeGen and Mooncake are slightly shifted for
%   visibility; their selected-prefix Jaccard scores are both 1.000.}
%   \label{fig:stability}
%   \vspace{-10pt}
% \end{figure}

All 81 trace-specific evaluations satisfy the direct two-witness gate.
BPW produces highly consistent replay selections across parameter,
configuration, and restart variations. BurstGPT shows mild ordering
sensitivity under the weakest source condition, requiring 4--5 workloads with
selected-prefix Jaccard values ranging from 0.667 to 1.000. In contrast,
ServeGen and Mooncake remain fully stable across all evaluated settings,
requiring only three workloads and achieving a selected-prefix Jaccard of
1.000. Across configuration variations, BPW maintains strong selection consistency,
with mean selected-prefix Jaccard values exceeding 0.93 on all three traces.
Restart evaluations show similarly stable behavior, indicating that the
selected replay suites are not artifacts of a particular parameter setting or
execution instance under diverse source and target conditions.
\section{Conclusion}
We propose Bottleneck-Preserving Witnessing (BPW), a compact diagnostic
replay framework for reliable LLM-serving evaluation. BPW combines
candidate nomination, coverage-priority sequencing, and direct
verification to preserve component-level bottleneck evidence. Experiments
on three public traces show that BPW achieves the best diagnostic quality
among 16 policies, improving Mean prefix Macro-F1 by 2.3 percent and
WBRC-AUC by 16.3 percent on average. Thus, BPW provides a compact and
reliable mechanism for bottleneck diagnosis while supporting cost-aware
LLM-serving evaluation. More broadly, BPW highlights that representative
workload selection is insufficient for mechanism-level diagnosis when
bottleneck evidence is non-compensatory. By shifting trace reduction from
distribution preservation toward evidence preservation, BPW offers a
general perspective for efficient and reliable serving-system evaluation
across diverse deployment settings.

\bibliographystyle{reference}
\bibliography{reference}

\end{document}